\documentclass[10pt,twocolumn,letterpaper]{article}

\usepackage{iccv}
\usepackage{times}
\usepackage{booktabs}
\usepackage{epsfig}
\usepackage{graphicx}
\usepackage{amsmath}
\usepackage{amssymb}
\usepackage{multirow}
\usepackage{multicol}
\usepackage{makecell}
\usepackage{subcaption}
\usepackage{bbm}
\usepackage{algorithm}
\usepackage[noend]{algorithmic}
\usepackage{array} 
\usepackage{setspace}
\usepackage[accsupp]{axessibility}  
\usepackage[breaklinks=true,bookmarks=false,colorlinks=true]{hyperref}

\usepackage[capitalize]{cleveref}
\crefname{section}{Sec.}{Secs.}
\Crefname{section}{Section}{Sections}
\Crefname{table}{Table}{Tables}
\crefname{table}{Tab.}{Tabs.}

\iccvfinalcopy 

\def\iccvPaperID{****} 

\ificcvfinal\fi

\begin{document}

\title{FRAug: Tackling Federated Learning with Non-IID Features via Representation Augmentation}

\author{Haokun Chen \textsuperscript{\rm 1,}\textsuperscript{\rm 2} \quad 
Ahmed Frikha \textsuperscript{\rm 1,}\textsuperscript{\rm 2,}\textsuperscript{\rm 3} \quad 
Denis Krompass \textsuperscript{\rm 2} \quad 
Jindong Gu \textsuperscript{\rm 4}\thanks{Corresponding author} \quad 
Volker Tresp \textsuperscript{\rm 1,}\textsuperscript{\rm 3} \\
\textsuperscript{\rm 1} Ludwig Maximilian University of Munich \quad  \textsuperscript{\rm 2} Siemens Technology \\
\textsuperscript{\rm 3} Munich Center for Machine Learning \quad \textsuperscript{\rm 4} University of Oxford\\
{\tt\small \{haokun.chen, ahmed.frikha, denis.krompass\}@siemens.com,} \\ 
{\tt\small jindong.gu@outlook.com, volker.tresp@lmu.de} \\
}

\maketitle
\ificcvfinal\thispagestyle{empty}\fi

\begin{abstract}
Federated Learning (FL) is a decentralized machine learning paradigm, in which multiple clients collaboratively train neural networks without centralizing their local data, and hence preserve data privacy. However, real-world FL applications usually encounter challenges arising from distribution shifts across the local datasets of individual clients. These shifts may drift the global model aggregation or result in convergence to deflected local optimum. While existing efforts have addressed distribution shifts in the label space, an equally important challenge remains relatively unexplored. This challenge involves situations where the local data of different clients indicate identical label distributions but exhibit divergent feature distributions. This issue can significantly impact the global model performance in the FL framework. In this work, we propose Federated Representation Augmentation (FRAug) to resolve this practical and challenging problem. FRAug optimizes a shared embedding generator to capture client consensus. Its output synthetic embeddings are transformed into client-specific by a locally optimized RTNet to augment the training space of each client. Our empirical evaluation on three public benchmarks and a real-world medical dataset demonstrates the effectiveness of the proposed method, which substantially outperforms the current state-of-the-art FL methods for feature distribution shifts, including PartialFed and FedBN. 
\end{abstract}

\section{Introduction}
\label{sec:intro}

Federated Learning (FL) is a machine learning paradigm in which a shared model is collaboratively trained using decentralized data sources. In the classical FL approach, \eg, FedAvg \cite{mcmahan2017communication}, the central server obtains the model by iteratively averaging the optimized model weights uploaded from the active clients. FL has the benefit that it does not require direct access to the client local datasets, resulting in improved client-server communication efficiency and enhanced data confidentiality.

Despite these promising prospects, real-world FL applications encounter practical challenges arising from data heterogeneity, in which the client local datasets are not independent and identically distributed (\emph{non-IID}). Non-IID data from different clients may cause local model drifts during the client update and overfitting to its local objective, making it challenging to obtain a stable and optimal convergence of the aggregated server model \cite{li2020federatedchallenge, mendieta2022local}.

As discussed in \cite{kairouz2021advances}, data heterogeneity in FL can be categorized into label space heterogeneity and feature space heterogeneity. A variety of methods were developed to tackle problem settings where the client datasets are non-IID in the label space \cite{zhu2021federated, zhang2021survey}. However, the under-explored problem of feature distribution shift is also prevalent in real-world applications, \eg, in the data collected from different scanners in clinical centers \cite{dou2019domain}, as well as gathered by different machines in industrial manufacturing plants \cite{li2020review}. Most importantly, although these entities may diagnose the same types of cancers or detect the same types of anomalies, \ie, having the same label distribution, they are not willing to share their original data to prevent competitive disadvantage or reverse engineering. Therefore, we propose an effective and privacy-preserving FL algorithm, \ie, Federated Representation Augmentation (\emph{FRAug}), to address this practical problem of feature space heterogeneity.

Unlike previous works that generate synthetic samples in the input space \cite{zhang2021fedzkt, zhang2022fine} or acquire additional public datasets \cite{lin2020ensemble, gong2022preserving}, FRAug applies data augmentation in the low-dimensional feature embedding space, which is more efficient and confronts fewer confidentiality threats. Moreover, the proposed augmentation algorithm is especially suitable for FL applications, where collaborative training is often conducted by multiple edge devices (clients) with limited computational powers and data quantities \cite{mcmahan2017communication}. Specifically, we first aggregate the consensual knowledge from different clients in the embedding space by training a shared representation generator, which produces client-agnostic embeddings. However, solely optimizing the generator might be challenging, given its training representations following different local client feature distributions. Therefore, a Representation Transformation Network (RTNet) is locally trained at each client to transform the client-agnostic synthetic embeddings into client-specific. Hereby, we aim at aligning the client-agnostic embeddings with the local feature distribution. Finally, the local dataset of each client will be augmented by its client-specific synthetic embeddings.


The proposed method FRAug achieves state-of-the-art results on three benchmark datasets with feature distribution shift, surpassing the concurrent FL methods addressing the same problem, including PartialFed \cite{sun2021partialfed} and FedBN \cite{li2021fedbn}. Moreover, the superior performance of FRAug on a medical dataset illustrates its applicability in complex real-world FL applications. Our contributions can be summarized as follows:

\begin{itemize}
    \item We propose a novel representation augmentation algorithm (\emph{FRAug}) to address FL with non-IID features.
    
    \item We conduct comprehensive experiments on three public benchmark datasets with feature distribution shifts, in which FRAug achieves SOTA results. 
    
    \item We verify the maturity and scalability of FRAug on a real-world medical dataset, and further analyze the convergence rate and robustness of FRAug.
    
    
\end{itemize}

\section{Related Work}
\subsection{Federated Learning (FL)} Federated Averaging (FedAvg) \cite{mcmahan2017communication} is one of the classic FL algorithms for training machine learning models using decentralized data sources. This simple paradigm suffers from performance degradation when there exists data heterogeneity \cite{kairouz2021advances, li2020federatedchallenge}. Numerous studies have been conducted for label space heterogeneity, \ie, class distributions are imbalanced across different clients, 
by adding additional regularization term in the client local update \cite{li2020federated, dandi2022implicit, ruan2022fedsoft, lee2021preservation, jeong2022factorized, chen2022calfat, kim2022multi, yu2021fed2}, utilizing shared local data \cite{zhao2018federated, liu2021feddg, gong2021ensemble}, introducing additional public datasets \cite{li2019fedmd, lin2020ensemble, gong2022preserving}, fully or partially personalizing the client models \cite{arivazhagan2019federated, fallah2020personalized, t2020personalized, li2021ditto, collins2021exploiting, oh2021fedbabu, alam2022fedrolex}, or performing data-free knowledge distillation \cite{lopes2017data} in the input space \cite{hao2021towards, zhang2021fedzkt, zhang2022fine} or the feature space \cite{he2020group, zhu2021data, luo2021no}. 
However, there are only limited studies addressing the heterogeneity in feature space, \ie, non-IID features. Recently, \cite{andreux2020siloed} showed that Batch Normalization layers (BN) \cite{ioffe2015batch} with local statistics improve the robustness of the FL model to inter-center data variability and yield better out-of-domain generalization results, while FedBN \cite{li2021fedbn} provided more theoretical analysis on the benefits of local BN layers for FL with feature non-IID. PartialFed \cite{sun2021partialfed} empirically found that partially initializing the client models could alleviate the effect of feature distribution shift. HarmoFL \cite{jiang2022harmofl} focused on FL applications for heterogeneous medical images and applied amplitude normalization in frequency space and model weight perturbation to harmonize the training process. In this work, we tackle the problem of non-IID features in FL via a client-specific data augmentation approach performed in the embedding space. In particular, client-agnostic embeddings are initially synthesized by a shared generator that captures the knowledge from different distributions, which are then personalized by separate client-specific models. Training the local model with the resulting client-specific embeddings improves its robustness against the feature distribution shift.

\subsection{Cross-Domain Learning} The problem of learning on centralized data with non-IID features, \ie, cross-domain data, has been widely studied in the context of Unsupervised Domain Adaptation (UDA) \cite{wilson2020survey, chen2022evidential, zhang2022interpretable, chu2022denoised, karisani2022multiple, xie2022gearnet}, where a model is trained using multiple source domains and finetuned using an unlabelled target domain, and Domain Generalization (DG) \cite{zhou2021domain, frikha2021towards, zhou2021domainvision, frikha2021columbus, li2022invariant, kang2022style}, where the target domain data is not accessible during the training process of UDA. A variety of efforts have been made to tackle the problem of UDA and DG. CROSSGRAD \cite{shankar2018generalizing} used adversarial gradients obtained from a domain classifier to augment the training data. L2A-OT \cite{zhou2020learning} trained a generative model to transfer the training samples into pseudo-novel domains. MixStyle \cite{zhou2021domainstyle} performed feature-level augmentation by interpolating the style statistics of the output features from different network layers. While the aforementioned methods assume centralized access to all datasets from different domains, we address the problem where the datasets are decentralized and cannot be shared due to privacy concerns.

\section{Methodology}
\subsection{Problem Statement}
In this work, we address an FL problem setting with non-IID features, which we describe in the following. Let $\mathcal{X} \subset \mathbb{R}^{d_{in}}$ be an input space, $\mathcal{U} \subset \mathbb{R}^{d_u}$ be a feature space, and $\mathcal{Y} \subset \mathbb{N}$ be an output space. Let $\boldsymbol{\theta}:=[\boldsymbol{\theta}_f, \boldsymbol{\theta}_h]$ denote the parameters of the classification model trained in an FL setting involving one central server and $K \in \mathbb{N}$ clients. The model consists of two components: a feature extractor $f: \mathcal{X} \to \mathcal{U}$ parameterized by $\boldsymbol{\theta}_f$, and a prediction head $h: \mathcal{U} \to \mathcal{Y}$ parameterized by $\boldsymbol{\theta}_h$. We assume that a dataset $D^k = \{(\boldsymbol{x}^k_i, y^k_i)|i \in \{1,..,N_k\}\}$, containing private data, is available on each client, where $N^k \in \mathbb{N}$ denotes the number of samples in $D^k$ and $C \in \mathbb{N}$ denotes the number of classes. As discussed in \cite{kairouz2021advances}, FL with non-IID data can be described by the distribution shift on local datasets: $P_{\mathcal{X}\mathcal{Y}}^{k_1} \neq P_{\mathcal{X}\mathcal{Y}}^{k_2}$ with $\forall k_1,k_2 \in \{1,...,K\}, k_1 \neq k_2$, where $P_{\mathcal{X}\mathcal{Y}}^k$ defines the joint distribution of input space $\mathcal{X}$ and label space $\mathcal{Y}$ on $D^k$. The addressed problem setting, \ie, FL with non-IID features, covers (1) \emph{covariate shift}: The marginal distribution $P_{\mathcal{X}}$ varies across clients, while $P_{\mathcal{Y}|\mathcal{X}}$ is the same, and (2) \emph{concept shift}: The conditional distribution $P_{\mathcal{X}|\mathcal{Y}}$ varies across clients, while $P_{\mathcal{Y}}$ is the same \cite{li2021fedbn}. From the perspective of cross-domain learning literature \cite{wilson2020survey, zhou2021domain}, local data from every client can be viewed as a separate domain.

\subsection{Motivational Case Study}
\label{sec:motivstudy}
To motivate our representation augmentation algorithm, we present an empirical analysis to address the following research question: \emph{Does finetuning only the prediction head using additional synthetic feature embeddings lead to performance improvement?} First, we optimize a classification model $\boldsymbol{\theta}^k$ with $10\%$ of the local dataset $D^k$ following prior FL work \cite{mcmahan2017communication, li2021fedbn}. Then, we fix the feature extractor and finetune \emph{only} the prediction head with $100\%$ of $D^k$. Finally, we evaluate both classification models. Here, we use the representations, extracted by the feature extractor using the additional real images, to simulate the output produced by a "perfect" embedding generator.

\begin{table}[t]
\centering
\small
\setlength{\tabcolsep}{1.2mm}{
\begin{tabular}{c|cccc|c}
\toprule
\multirow{2}{*}{Method} & \multicolumn{5}{c}{OfficeHome}   \\
\cline{2-6}
~ & Art & Clipart & Product & Real & avg \\
\hline
w/o Add. Embeddings & 57.47 & 56.74 & 73.32 & 71.25 & 64.69 \\
w. Add. Embeddings & 68.18 & 72.31 & 80.04 & 79.50 & \textbf{75.01} \\
\bottomrule
\end{tabular}
\vspace{-0.3em}
\caption{Evaluation accuracies of models optimized with (w.) and without (w/o) prediction head finetuned using additional embeddings on OfficeHome benchmark, indicating the applicability of the representation generator given the performance increase.}
\label{tab:motiv}
}
\vspace{-0.6em}
\end{table}

The results in \cref{tab:motiv} show that the feature extractor, trained with less data, still captures useful information when exposed to unseen image samples. Most importantly, a substantial average performance boost of $10.32\%$ shows that generating additional representations benefits the client local update, proving the applicability and effectiveness of the proposed method.

\subsection{Proposed Method}
\label{sec:FRAug}
To tackle FL with non-IID features, we propose Federated Representation Augmentation (FRAug). Our algorithm is built upon FedAvg \cite{mcmahan2017communication}, which is the most widely used FL strategy. In FedAvg, the central server sends a copy of the global model $\boldsymbol{\theta}$ to each client to initialize their local models $\{\boldsymbol{\theta}^k| k \in K\}$. After training on its local dataset $D^k$, the client-specific updated models are sent back to the central server, where they are averaged and used as the global model. Such communication rounds are repeated until some predefined convergence criteria are met. Similarly, the training process of FRAug (\cref{algo:FRAug}) can be divided into two stages: (1) The \emph{Server Update}, where the central server aggregates the parameters uploaded by the clients and distributes the averaged parameters to each client, and (2) the \emph{Client Update}, where each client receives the model parameters from the central server and performs local optimization. Unlike FedAvg, where only the local dataset of each client is used for training, FRAug generates additional feature embeddings to finetune the prediction head of the local classification model. Concretely, we train a shared generator and a local Representation Transformation Network (RTNet) for each client, which together produce \emph{domain-specific} synthetic feature embeddings for each client to augment its local data in the embedding space. Hereby, the shared generator captures knowledge from all the clients to generate client-agnostic embeddings, which are then personalized by the local RTNet into client-specific embeddings. In the following, we provide a more detailed explanation of FRAug.

\begin{figure}[t]
  \hspace{-0.2cm}\includegraphics[scale=0.34]{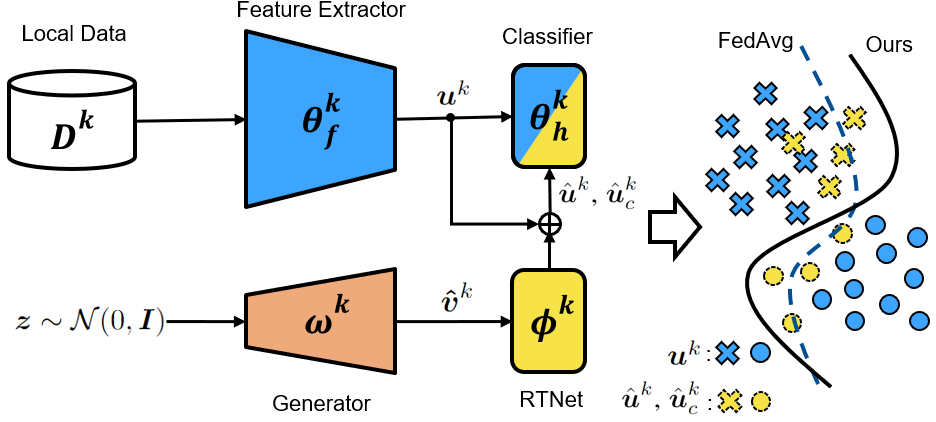}
  \raggedleft
  \caption{Overview of FRAug local update at client $k$: a shared generator is learned to aggregate knowledge from multiple clients and generate client-agnostic feature embeddings $\hat{\boldsymbol{v}}^k$, which are then fed into the local Representation Transformation Network (\emph{RTNet}) to produce client-specific feature embeddings $\hat{\boldsymbol{u}}^k$ and $\hat{\boldsymbol{u}}^k_c$. Finally, the real feature embeddings $\boldsymbol{u}^k$, extracted by the feature extractor using local dataset $D^k$, will be augmented with $\hat{\boldsymbol{u}}^k$ and $\hat{\boldsymbol{u}}^k_c$ in the classification model optimization.}
  \label{fig:overviewFRAug}
  \vspace{-10pt}
\end{figure}

\subsubsection{Server Update} At the beginning of the training, the server initializes the parameters of the classification model $\boldsymbol{\theta}:=[\boldsymbol{\theta}_f, \boldsymbol{\theta}_h]$, as well as the \emph{shared} generator $\boldsymbol{\omega}$. In each communication round $r$, all clients receive the aggregated model parameters and conduct the \emph{Client Update} procedure in parallel. Subsequently, the server securely aggregates the optimized model parameters from all the clients into a single model that is used in the next communication round.

\subsubsection{Client Update} 
\label{sec:clientupdate}
As shown in \cref{fig:overviewFRAug}, at the beginning of the first communication round, each client \emph{locally} initializes a Representation Transformation Network (\emph{RTNet}) parameterized by $\boldsymbol{\phi}^k$. Subsequently, each client receives the classification model parameters $\boldsymbol{\theta}^k$ and the generator parameters $\boldsymbol{\omega}^k$ from the server, and conducts $T$ local update steps. Each local update comprises 2 stages: (1) Classification model optimization, and (2) Generator and RTNet optimization.

\textbf{(1) Classification Model Optimization:} In this stage, the generator and the RTNet are fixed, while the classification model is updated by minimizing the loss $\mathcal{L}_{cls}$, where 

\vspace{-0.4cm}
\begin{equation} 
\begin{split}
\label{eq:Lcls}
    &\mathcal{L}_{cls} = \mathcal{L}_{real} + \mathcal{L}_{syn},\\
    &\text{with} \quad \mathcal{L}_{real} = L_{\text{CE}}(h^k(f^k(\boldsymbol{X}^k)), \boldsymbol{y}^k).
\end{split}
\end{equation}

While $\mathcal{L}_{real}$ is minimized to update the model parameter $\boldsymbol{\theta}^k$ by using real training samples from $D^k$, $\mathcal{L}_{syn}$ is minimized to update only the prediction head $h^k$ as it is computed on synthetically generated samples in the embedding space $\mathcal{U}$. We use cross-entropy ($L_{CE}$) for both loss functions.

To generate domain-specific synthetic embeddings, the shared generator $g^k$ and local RTNet $m^k$ are used to generate residuals that are added to the embeddings of real examples produced by the local feature extractor $f^k$. Hereby, we first generate client-agnostic embeddings $\boldsymbol{\hat{v}}^k$ by feeding a batch of random vector $\boldsymbol{z}$, sampled from standard Gaussian distribution $\mathcal{N}(0, \boldsymbol{I})$, and class labels $\boldsymbol{y}$ into the generator $g^k$. Subsequently, $\boldsymbol{\hat{v}}^k$ are transformed by the local RTNet into client-specific residuals and added to the embeddings of real datapoints. We distinguish two types of synthetic embeddings that we generate to train the local prediction head: domain-specific synthetic embeddings $\hat{\boldsymbol{u}}^{k}$ and class-prototypical domain-specific synthetic embeddings $\hat{\boldsymbol{u}}^{k}_c$ for category $c$. The domain-specific embeddings $\hat{\boldsymbol{u}}^{k}$ are generated by adding synthetic residuals to the embeddings $\boldsymbol{u}^k$ of real examples from the current batch sampled from $D^k$. On the other hand, synthetic residuals are added to class-prototypes $\overline{\boldsymbol{u}}^k_c$, \ie, class-wise average embeddings of real examples, to produce $\hat{\boldsymbol{u}}^{k}_c$, which stabilizes the training and increase the variance of the generated embeddings.

\vspace{-0.4cm}
\begin{equation} 
\label{eq:Lsyn}
    \mathcal{L}_{syn}  = L_{\text{CE}}(h^k(\hat{\boldsymbol{u}}^{k}), \boldsymbol{y}) + \sum_{c \in C}L_{\text{CE}}(h^k(\hat{\boldsymbol{u}}^{k}_c), c),
\end{equation} 

\vspace{-0.7cm}
\begin{equation} 
\begin{split}
\label{eq:featuredef}
    \text{with} \quad & \hat{\boldsymbol{u}}^{k} = \boldsymbol{u}^k + \lambda_{syn} \cdot m^k(g^k(\boldsymbol{z}, \boldsymbol{y})), \\
    &\hat{\boldsymbol{u}}^{k}_c = \overline{\boldsymbol{u}}^k_c + \lambda_{syn} \cdot m^k(g^k(\boldsymbol{z}^\prime, c)).
\end{split}
\end{equation} 
\vspace{-0.2cm}

To compute the class-wise average embedding $\overline{\boldsymbol{u}}^k_c$, we use the exponential moving average (EMA) scheme, at each local iteration. In particular,

\vspace{-0.4cm}
\begin{equation} 
\label{eq:computemean}
    \overline{\boldsymbol{u}}^k_c \leftarrow 
         (1-\lambda_c)\cdot {\overline{\boldsymbol{u}}^k_c} + \lambda_c \cdot \frac{\sum_{i \in B} \mathbbm{1}(\boldsymbol{y}_i=c) \cdot f(\boldsymbol{x}_i)}{\sum_{i \in B} \mathbbm{1}(\boldsymbol{y}_i=c) + \epsilon}, 
\end{equation}
\vspace{-0.3cm}

\noindent
where $\mathbbm{1}(\cdot)$ denotes the indicator function, $B$ is the batch size of the real samples, and $\epsilon$ is a small number added for numerical stability. By using the average embeddings of previous iterations, we enable the examples of previously sampled batches to contribute to the computation of the current average embeddings. The ratio $\lambda_c$ follows an exponential ramp-up schedule as proposed in \cite{laine2016temporal}. 

We note that, in \cref{eq:featuredef}, for the generation of $\hat{\boldsymbol{u}}^{k}$, the original labels $\boldsymbol{y}$ of the sampled data batch are used for the residual generation, since the residuals are added to the embeddings of the examples corresponding to these labels. For $\hat{\boldsymbol{u}}^{k}_c$, we feed the label $c$ that corresponds to the class of the average embedding $\overline{\boldsymbol{u}}^k_c$. While the residuals produced by the generator and the RTNet are random in early training iterations due to the random initialization of these models, they become more informative as training progresses. To reflect this in our algorithm, we employ the weighting coefficient $\lambda_{syn}$ (\cref{eq:featuredef}) that controls the impact of the residuals, and increase it following an exponential schedule throughout training. 

To allow the different client-specific models to learn feature extractors tailored to their data distribution $D^k$, while still benefiting from the collaborative learning, we use local Batch Normalization layers (BN) \cite{ioffe2015batch} as introduced in \cite{li2021fedbn}.

\textbf{(2) Generator and RTNet Optimization:} In the second stage, the classification model is fixed while the generator and the RTNet are optimized. The class-conditional generator $g^k$ takes a batch of random vectors $\boldsymbol{z}$ and class labels $\boldsymbol{y}$ to produce \emph{client-agnostic} feature embeddings $\hat{\boldsymbol{v}}^k$. $\hat{\boldsymbol{v}}^k$ are then fed into the RTNet $m^k$ to be adapted to the feature distribution of the corresponding client $k$. The resulting residuals are added on the embeddings of real examples to produce the \emph{domain-specific} synthetic embeddings $\hat{\boldsymbol{u}}^k$ and $\hat{\boldsymbol{u}}^k_c$. The generator will be optimized by minimizing the loss $\mathcal{L}_{gen}$, with 

\vspace{-0.6cm}
\begin{equation}
\label{eq:Lgenerator}
    \mathcal{L}_{gen} = L_{\text{CE}}(h^k(\hat{\boldsymbol{v}}^k), \boldsymbol{y}) - \alpha L_{\text{MMD}}(\hat{\boldsymbol{v}}^k, \boldsymbol{u}^k).
\end{equation}
\vspace{-0.6cm}

The minimization of the cross-entropy loss $L_{\text{CE}}$ incentivizes the shared generator to produce features that are recognized by the prediction heads of all the clients. By sharing and optimizing the generator across all clients, we ensure that the synthetic embeddings produced by the generator, \ie, $\hat{\boldsymbol{v}}^k$, capture client-agnostic semantic information. Additionally, we maximize the statistical distance \cite{wootters1981statistical} between $\hat{\boldsymbol{v}}^k$ and the real feature embeddings $\boldsymbol{u}^k$. By doing so, we force $\hat{\boldsymbol{v}}^k$ not to follow any client-specific distribution, and thus enhance the variance of the augmented feature space. Here, we adopt Maximum Mean Discrepancy (MMD) \cite{gretton2012kernel} as the distance metric. Subsequently, the client-agnostic embeddings are fed into the RTNet $m^k$ parametrized by $\boldsymbol{\phi}^k$ to produce domain-specific embeddings $\hat{\boldsymbol{u}}^k$ and $\hat{\boldsymbol{u}}^k_c$. $\boldsymbol{\phi}^k$ is optimized by minimizing the loss $\mathcal{L}_{rt}$, where

\begin{equation}
\vspace{-5pt}
\begin{aligned}
\label{eq:LRTNet}
    \mathcal{L}_{rt} = &-L_{\text{ent}}(h^k(\hat{\boldsymbol{u}}^k)) - \sum_{c \in C} L_{\text{ent}}(h^k(\hat{\boldsymbol{u}}^k_c)) \\ &+\beta  (L_{\text{MMD}}(\hat{\boldsymbol{u}}^k, \boldsymbol{u}^k) + \sum_{c \in C} L_{\text{MMD}}(\hat{\boldsymbol{u}}^k_c, \overline{\boldsymbol{u}}^k_c)).
\end{aligned}
\end{equation}

Here, we maximize the entropy ($L_{\text{ent}}$) of the prediction head output on $\hat{\boldsymbol{u}}^k$, $\hat{\boldsymbol{u}}^k_c$ to incentivize the generation of synthetic embeddings that are \emph{hard} to classify for the prediction head $h^k$. To avoid generating outliers, we align the synthetic embedding distribution with that of the client local data by minimizing their Maximum Mean Discrepancy (MMD). In particular, we penalize high MMD distances between $\hat{\boldsymbol{u}}^k$ and $\boldsymbol{u}^k$, as well as $\hat{\boldsymbol{u}}^k_c$ and $\overline{\boldsymbol{u}}^k_c$ for each class $c$. $\alpha$ and $\beta$ denote weighting coefficients in \cref{eq:Lgenerator} and \cref{eq:LRTNet}, respectively.

\begin{algorithm}[tb]
\small
\caption{Training procedure of FRAug}\label{algo:FRAug}
\textbf{ServerUpdate}
\begin{algorithmic}[1]
\STATE Randomly initialize $\boldsymbol{\theta}_0, \boldsymbol{\omega}_0$ \\
\FOR {round $r$ = 1 to $R$} 
    \FOR [\textbf{in parallel}]{client $k$ = 1 to $K$} 
        \STATE $\boldsymbol{\theta}^k_r, \boldsymbol{\omega}^k_r \gets$ ClientUpdate$(\boldsymbol{\theta}_{r-1}, \boldsymbol{\omega}_{r-1}, k, r)$
    \ENDFOR
    \STATE$\boldsymbol{\theta}_{r} \gets \frac{1}{K} \sum_{k=1}^K  \boldsymbol{\theta}^k_{r}$ \\
    \STATE$\boldsymbol{\omega}_{r} \gets \frac{1}{K} \sum_{k=1}^K  \boldsymbol{\omega}^k_{r}$\\
\ENDFOR
\vspace{0.5em}
\end{algorithmic} 
\textbf{ClientUpdate}$(\boldsymbol{\theta}, \boldsymbol{\omega}, k, r)$
\begin{algorithmic}[1]
\IF {$r=1$}
    \STATE Randomly initialize $\boldsymbol{\phi}^k$ \\
\ENDIF
\STATE $\boldsymbol{\theta}^k \gets \boldsymbol{\theta}$, \ $\boldsymbol{\omega}^k \gets \boldsymbol{\omega}$ \\
\FOR {local step $t$ = 1 to $T$} 
    \STATE Sample $\{\boldsymbol{X}, \boldsymbol{y}\}$ from $D_k$ \\
    \STATE Sample $ \boldsymbol{z}, \boldsymbol{z}^{\prime} \sim \mathcal{N}(0, I) $\\
    \STATE Optimize $\boldsymbol{\theta}^k$ (\cref{eq:Lcls})\\ 
    \STATE Optimize $\boldsymbol{\omega}^k$ (\cref{eq:Lgenerator}) and $\boldsymbol{\phi}^k$ (\cref{eq:LRTNet})\\
\ENDFOR
\end{algorithmic}
\end{algorithm}

\section{Experiments and Analyses}
We conduct an extensive empirical analysis to investigate the proposed method and its viability. Firstly, we compare FRAug with several FL baseline methods on 3 popular benchmark datasets involving feature distribution shifts. Subsequently, we validate our approach on a real-world medical dataset for genetic treatment classification. We present additional analysis regarding convergence rate, communication overhead, and robustness to input noise. Finally, we demonstrate the ablation studies of FRAug and its comparison with other augmentation-based FL methods. 

\subsection{Benchmark Experiments}
\label{sec:dataset}
\subsubsection{Datasets Description} We conduct experiments on three common image classification benchmarks with domain shift: (1) \emph{OfficeHome} \cite{venkateswara2017deep}, which contains 65 classes in four domains: Art (A), Clipart (C), Product (P) and Real-World (R). (2) \emph{PACS} \cite{li2017deeper}, which includes images that belong to 7 classes from four domains Art-Painting (A), Cartoon (C), Photo (P), and Sketch (S). (3) \emph{Digits} comprises images of 10 digits from the following four datasets: MNIST (MT) \cite{lecun1998gradient}, MNIST-M (MM) \cite{ganin2015unsupervised}, SVHN (SV) \cite{netzer2011reading}, and USPS (UP) \cite{hull1994database}. Each client contains data from one of the domains, \ie, there exists feature distribution shifts across different clients. To simulate data scarcity described in previous sections, we assume that only $10\%$ ($1\%$ for the Digits dataset) of the original data is available for each client, resulting in ca. 100 to 1000 data samples per client following the experimental setup in the previous work \cite{mcmahan2017communication, li2021fedbn}.

\subsubsection{Baselines} We compare our approach with several baseline methods, including \emph{Single}, \ie, training an individual model on each client separately, \emph{All}, \ie, training a single model at the central server using data aggregated from all clients, \emph{FedAvg} \cite{mcmahan2017communication}, \emph{pFedAvg}, \ie, FedAvg with local model personalization. We also compare FRAug with \emph{FedProx} \cite{li2020federated}, \emph{FedBABU} \cite{oh2021fedbabu}, and \emph{FedProto} \cite{tan2022fedproto}, which are strong concurrent methods handling label space heterogeneity in FL. We note that \emph{All} is an oracle baseline as it requires centralizing the data from the different clients, hence infringing the data-privacy requirements. Furthermore, we compare our method with the current state-of-the-art FL methods for non-IID features, \ie, \emph{FedBN} \cite{li2021fedbn} and \emph{PartialFed} \cite{sun2021partialfed}. We use the published code of \emph{FedBN} and reimplement \emph{PartialFed} since the original implementation was not made public. We conduct the same hyperparameter search for all methods and report the best results. The detailed hyperparameter search spaces of different methods are provided in Appendix A. 

\subsubsection{Implementation Details} For the OfficeHome and PACS datasets, we use a ResNet18 \cite{he2016deep} pretrained on ImageNet \cite{deng2009imagenet} as initialization of the classification model. For Digits, we use a 6-layer Convolution Neural Network (CNN) as the backbone following prior work \cite{li2021fedbn}. We adopt a 2-layer MLP as the generator and RTNet architectures for all datasets. Besides, we apply the same data augmentation techniques on the input images during the classification model training for all clients following the previous work  \cite{gulrajani2020search}. In Appendix A, we provide further details about model architectures and training hyperparameters. All experiments are repeated with 3 different random seeds. 

\begin{table*}[ht]
\begin{center}
\small
\centering
\setlength\tabcolsep{5pt}
\begin{tabular}{cc|cccccccccc}

\toprule
\multicolumn{2}{c|}{Benchmark}                                                    & Single                  & All                     & FedAvg                   & FedProx                 
        & FedProto                 & FedBABU    & pFedAvg                 & PartialFed               & FedBN                    & FRAug                   \\

\hline
\multirow{5}{*}{\makecell[c]{Office\\Home}}
& A   & 35.80{\scriptsize ±0.1} & 56.65{\scriptsize ±0.7} & 57.47{\scriptsize ±0.6} & 55.68{\scriptsize ±0.4} & 51.44{\scriptsize ±0.6} & 49.80{\scriptsize ±0.4} & 52.50{\scriptsize ±0.9} & 48.83{\scriptsize ±0.2} & 57.59{\scriptsize ±0.8} & \textbf{57.61}{\scriptsize ±0.6} \\
& C   & 45.54{\scriptsize ±0.8} & 58.81{\scriptsize ±1.6} & 56.74{\scriptsize ±0.9} & 56.88{\scriptsize ±0.5} & 52.63{\scriptsize ±0.7} & 54.23{\scriptsize ±0.7} & 52.09{\scriptsize ±1.1} & 49.96{\scriptsize ±0.2}  & 56.52{\scriptsize ±0.3}  & \textbf{60.03}{\scriptsize ±0.5} \\
& P   & 67.04{\scriptsize ±0.8} & 71.39{\scriptsize ±0.3} & 73.32{\scriptsize ±0.8} & 73.84{\scriptsize ±0.3} & 70.78{\scriptsize ±0.7} & 70.72{\scriptsize ±0.6} & 71.78{\scriptsize ±0.8} & 72.22{\scriptsize ±0.8}  & 73.55{\scriptsize ±1.0}  & \textbf{74.03}{\scriptsize ±0.8} \\
& R   & 61.16{\scriptsize ±0.7} & 72.63{\scriptsize ±1.3} & 71.25{\scriptsize ±0.3} & 72.15{\scriptsize ±0.9} & 65.13{\scriptsize ±0.2} & 66.74{\scriptsize ±0.5} & 66.28{\scriptsize ±0.4} & 65.82{\scriptsize ±0.6}  & 72.40{\scriptsize ±0.9}  & \textbf{74.58}{\scriptsize ±0.4} \\
\cline{2-12}
& avg & 52.42{\scriptsize ±0.4} & 64.87{\scriptsize ±0.9} & 64.69{\scriptsize ±0.6}  & 64.63{\scriptsize ±0.6} & 60.00{\scriptsize ±0.3} & 60.37{\scriptsize ±0.3} & 60.67{\scriptsize ±0.7} & 59.20{\scriptsize ±0.5}  & 65.02{\scriptsize ±0.7}  & \textbf{66.60}{\scriptsize ±0.3} \\

\hline
\multirow{5}{*}{Digits} 
& MT  & 96.68{\scriptsize ±0.2} & 97.04{\scriptsize ±0.1} & 96.85{\scriptsize ±0.1}  & 96.90{\scriptsize ±0.1} & 96.80{\scriptsize ±0.1} & 97.38{\scriptsize ±0.2} & 96.40{\scriptsize ±0.2} & 97.13{\scriptsize ±0.1}  & 97.03{\scriptsize ±0.1}  & \textbf{97.81}{\scriptsize ±0.1} \\
& MM  & 77.77{\scriptsize ±0.5} & 77.04{\scriptsize ±0.1} & 73.51{\scriptsize ±0.2}  & 72.60{\scriptsize ±0.4} & 78.16{\scriptsize ±0.6} & 79.30{\scriptsize ±0.8} & 77.56{\scriptsize ±0.4} & 74.21{\scriptsize ±0.5}  & 77.02{\scriptsize ±0.2}  & \textbf{81.65}{\scriptsize ±0.5} \\
& SV  & 75.55{\scriptsize ±0.3} & 77.96{\scriptsize ±0.5} & 74.49{\scriptsize ±0.2}  & 73.01{\scriptsize ±0.5} & 77.90{\scriptsize ±0.2} & 74.03{\scriptsize ±0.5} & 77.50{\scriptsize ±0.1} & 78.10{\scriptsize ±0.5}  & 77.59{\scriptsize ±0.1}  & \textbf{81.24}{\scriptsize ±0.3} \\
& UP  & 79.93{\scriptsize ±0.8} & 97.13{\scriptsize ±0.1} & 97.62{\scriptsize ±0.1}  & 97.31{\scriptsize ±0.3} & 97.37{\scriptsize ±0.1} & 95.37{\scriptsize ±0.4} & 96.67{\scriptsize ±0.1} & 94.78{\scriptsize ±0.5}  & 96.80{\scriptsize ±0.2}  & \textbf{97.67}{\scriptsize ±0.3} \\
\cline{2-12}
 & avg & 82.54{\scriptsize ±0.1} & 87.29{\scriptsize ±0.2} & 85.62{\scriptsize ±0.2}  & 84.96{\scriptsize ±0.3} & 87.50{\scriptsize ±0.1} & 86.52{\scriptsize ±0.4} & 87.03{\scriptsize ±0.2} & 86.05{\scriptsize ±0.3}  & 87.11{\scriptsize ±0.2}  & \textbf{89.59}{\scriptsize ±0.4} \\

\hline
\multirow{5}{*}{PACS}   
& A   & 82.37{\scriptsize ±0.6} & 83.17{\scriptsize ±0.2} & 82.72{\scriptsize ±0.4}  & 80.17{\scriptsize ±0.4} & 85.09{\scriptsize ±0.5} & 81.25{\scriptsize ±0.6} & 88.05{\scriptsize ±0.8} & 84.85{\scriptsize ±0.2}  & 86.60{\scriptsize ±0.5}  & \textbf{87.34}{\scriptsize ±0.5} \\
& C   & 86.08{\scriptsize ±0.9} & 86.92{\scriptsize ±0.8} & 84.04{\scriptsize ±1.3}  & 82.04{\scriptsize ±0.8} & 86.91{\scriptsize ±0.3} & 87.76{\scriptsize ±1.1} & 86.20{\scriptsize ±0.7} & 87.92{\scriptsize ±0.5}  & 87.76{\scriptsize ±1.0}  & \textbf{88.47}{\scriptsize ±0.9} \\
& P   & 92.01{\scriptsize ±1.1} & 95.95{\scriptsize ±0.8} & 96.05{\scriptsize ±0.5}  & 96.74{\scriptsize ±1.0} & 96.49{\scriptsize ±0.6} & 94.74{\scriptsize ±0.4} & 97.89{\scriptsize ±0.5} & 98.24{\scriptsize ±0.4}  & 97.95{\scriptsize ±0.4}  & \textbf{98.64}{\scriptsize ±0.6} \\
& S   & 87.52{\scriptsize ±0.8} & 88.70{\scriptsize ±0.7} & 89.50{\scriptsize ±0.7}  & 88.50{\scriptsize ±1.0} & 89.20{\scriptsize ±0.4} & 89.41{\scriptsize ±0.3} & 88.89{\scriptsize ±0.9} & 90.10{\scriptsize ±0.8}  & 90.75{\scriptsize ±0.3}  & \textbf{90.95}{\scriptsize ±0.4} \\
\cline{2-12}
& avg & 87.00{\scriptsize ±0.5} & 88.68{\scriptsize ±0.6} & 88.08{\scriptsize ±0.9}  & 86.86{\scriptsize ±0.9} & 89.42{\scriptsize ±0.5} & 88.29{\scriptsize ±0.6} & 90.26{\scriptsize ±0.6} & 90.28{\scriptsize ±0.7}  & 90.76{\scriptsize ±0.3}  & \textbf{91.34}{\scriptsize ±0.1} \\
\bottomrule       
\end{tabular}
\end{center}
\vspace{-1.0em}
\caption{Evaluation results of different algorithms on three real-world benchmark datasets with feature distribution shift. We report the mean{\scriptsize±std} accuracy of each client from 3 runs with different seeds. The best results are marked in \textbf{bold} (The same applies to the subsequent tables).}
\label{tab:mainresults}
\vspace{-1em}
\end{table*}

\subsubsection{Results and Discussion} We report the accuracies achieved by the different methods on all three datasets in \cref{tab:mainresults}. We observe that FRAug outperforms all the baselines on all benchmark datasets. On OfficeHome, FRAug outperforms FedAvg and FedBN by $1.91\%$ and $1.58\%$, respectively. On Digits, FRAug achieves a substantial $2.3\%$ improvement on average compared with all the alternative methods. Likewise, FRAug yields the highest average accuracy on PACS. We note that FRAug achieves an average performance increase of 1.6\% compared to FedBN across all three datasets, which surpasses the average performance improvement yielded by FedBN on FedAvg, \ie, 1.5\%. Moreover, we find that the performance improvement compared to the best baseline is the highest on the most challenging domains, \ie, on which all methods yield lower results than on other domains. These include MNIST-M and SVHN from Digits, as well as Clipart from OfficeHome, where FRAug achieves impressive improvements of above $3\%$. Interestingly, our approach outperforms the centralized baseline \emph{All}, demonstrating its effectiveness in aggregating the knowledge from different clients to enable a client-specific augmentation. 

\subsection{Validation on a Real-World Medical Dataset} 
\begin{figure}[ht]
\vspace{-0.5em}
\begin{subfigure}[b]{0.12\textwidth}
    \centering
    \includegraphics[scale=0.205]{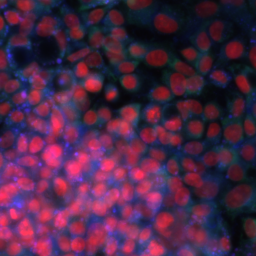} 
    \caption{HEPG2} 
\end{subfigure}%
\begin{subfigure}[b]{0.122\textwidth}
    \centering
    \includegraphics[scale=0.205]{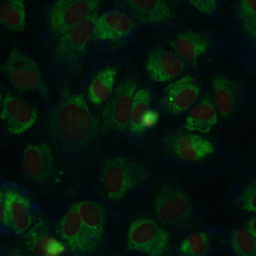} 
    \caption{HUVEC} 
\end{subfigure}%
\begin{subfigure}[b]{0.122\textwidth}
    \centering
    \includegraphics[scale=0.205]{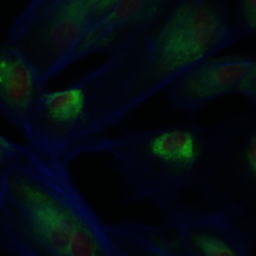} 
    \caption{RPE} 
\end{subfigure}%
\begin{subfigure}[b]{0.122\textwidth}
    \centering
    \includegraphics[scale=0.205]{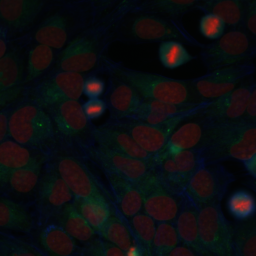} 
    \caption{U2OS} 
\end{subfigure}%
\vspace{-0.3em}
\caption{Example images of different cell types, \ie, local data from different clients, in RxRx1 dataset. Strong feature space heterogeneity can be observed between image appearance. \it{Best viewed in color.}}
\label{fig:ExampleRXRX1} 
\vspace{-1em}
\end{figure}

\subsubsection{Experimental Setup} To illustrate the effectiveness of FRAug on real-world applications, we further conduct experiments on the RxRx1 \cite{taylor2019rxrx1} medical dataset, which contains images (\cref{fig:ExampleRXRX1}) of cells obtained by fluorescent microscopy. The task is to classify which genetic treatment the cells received. There are 4 different cell types adopted in the dataset, \ie, HEPG2 (H), HUVEC (V), RPE (R), and U2OS (U), while multiple batches of experiments are executed for each cell type. Despite the careful control of experimental variables, \eg, temperature and humidity, feature space heterogeneity is observed across different batches of experiments \cite{koh2021wilds}. Therefore, we consider 4 different cell types as 4 different domains. We divide the batches of experiments from each domain, \ie, for each cell type, into 4 groups, where each group has the same number of batches and is assigned to one client. By doing so, we simulate a real-world collaborative training setup of different medical institutions where every institution has conducted some batches of experiments on one specific cell type. We note that the number of domains is not equal to the number of clients. Following the FL setting described in the previous section, we select 50 classes from 1139 classes in the original dataset. We adopt ResNet18 \cite{he2016deep} pretrained on ImageNet \cite{deng2009imagenet} as initialization of the classification model. To further evaluate the scalability of the proposed method, we conduct experiments where 2, 3, and 4 clients from each domain are selected, which gives in total 8, 12, and 16 clients joining the federated communication, respectively. Note that more clients correspond to larger data quantity.

\begin{table*}[htb]
\small
\setlength\tabcolsep{4.85pt}
\begin{tabular}{c|cccc|c|cccc|c|cccc|c}
\toprule
\multirow{2}{*}{Method} & \multicolumn{5}{c|}{8 clients} & \multicolumn{5}{c|}{12 clients} & \multicolumn{5}{c}{16 clients} \\
\cline{2-16}
~ & H  & V  & P  & U  & avg & H  & V  & P  & U  & avg & H  & V  & P  & U  & avg \\
\hline
FedAvg & \makecell[c]{24.31\\\scriptsize ±0.3} & \makecell[c]{34.39\\\scriptsize ±0.8} & \makecell[c]{20.19\\\scriptsize ±1.3} & \makecell[c]{17.65\\\scriptsize ±0.9} & \makecell[c]{24.14\\\scriptsize ±0.8} & \makecell[c]{28.84\\\scriptsize ±1.3} & \makecell[c]{40.60\\\scriptsize ±0.9} & \makecell[c]{19.72\\\scriptsize ±0.7} & \makecell[c]{16.67\\\scriptsize ±0.8} & \makecell[c]{26.46\\\scriptsize ±0.8} & \makecell[c]{28.17\\\scriptsize ±0.7} & \makecell[c]{41.60\\\scriptsize ±1.0} & \makecell[c]{23.55\\\scriptsize ±0.8} & \makecell[c]{17.65\\\scriptsize ±0.8} & \makecell[c]{27.74\\\scriptsize ±0.6} \\
\hline
HarmoFL & \makecell[c]{19.61\\\scriptsize ±1.0} & \makecell[c]{44.02\\\scriptsize ±0.5} & \makecell[c]{20.18\\\scriptsize ±0.2} & \makecell[c]{\textbf{22.53}\\\scriptsize ±0.9} & \makecell[c]{26.58\\\scriptsize ±1.0} & 
\makecell[c]{26.61\\\scriptsize ±0.8} & \makecell[c]{\textbf{49.15}\\\scriptsize ±0.5} & \makecell[c]{19.27\\\scriptsize ±0.7} & \makecell[c]{17.97\\\scriptsize ±0.9} & \makecell[c]{28.25\\\scriptsize ±0.8} & 
\makecell[c]{28.57\\\scriptsize ±0.9} & \makecell[c]{47.29\\\scriptsize ±0.7} & \makecell[c]{22.02\\\scriptsize ±0.5} & \makecell[c]{18.05\\\scriptsize ±0.7} & \makecell[c]{28.98\\\scriptsize ±0.4} \\
\hline
FedBN & \makecell[c]{22.94\\\scriptsize ±0.9} & \makecell[c]{43.70\\\scriptsize ±0.5} & \makecell[c]{25.92\\\scriptsize ±1.0} & \makecell[c]{18.63\\\scriptsize ±0.9} & \makecell[c]{27.80\\\scriptsize ±1.0} & \makecell[c]{27.22\\\scriptsize ±0.4} & \makecell[c]{46.01\\\scriptsize ±0.4} & \makecell[c]{26.85\\\scriptsize ±0.8} & \makecell[c]{16.95\\\scriptsize ±1.1} & \makecell[c]{29.26\\\scriptsize ±0.6} & \makecell[c]{29.35\\\scriptsize ±0.6} & \makecell[c]{\textbf{49.08}\\\scriptsize ±0.8} & \makecell[c]{29.58\\\scriptsize ±0.3} & \makecell[c]{19.97\\\scriptsize ±0.2} & \makecell[c]{31.99\\\scriptsize ±0.3} \\
\hline
FRAug & \makecell[c]{\textbf{28.28}\\\scriptsize ±0.3} & \makecell[c]{\textbf{45.33}\\\scriptsize ±0.9} & \makecell[c]{\textbf{28.74}\\\scriptsize ±1.2} & \makecell[c]{21.04\\\scriptsize ±0.5} & \makecell[c]{\textbf{30.84}\\\scriptsize ±0.5} & \makecell[c]{\textbf{30.73}\\\scriptsize ±0.9} & \makecell[c]{47.36\\\scriptsize ±0.8} & \makecell[c]{\textbf{30.58}\\\scriptsize ±0.2} & \makecell[c]{\textbf{19.60}\\\scriptsize ±0.7} & \makecell[c]{\textbf{32.07}\\\scriptsize ±0.5} & \makecell[c]{\textbf{32.34}\\\scriptsize ±0.4} & \makecell[c]{48.05\\\scriptsize ±0.5} & \makecell[c]{\textbf{31.83}\\\scriptsize ±1.0} & \makecell[c]{\textbf{20.59}\\\scriptsize ±0.7} & \makecell[c]{\textbf{33.20}\\\scriptsize ±0.8} \\ 
\bottomrule       
\end{tabular}
\vspace{-0.5em}
\caption{Evaluation results of different methods on real-world medical dataset RxRx1. We conduct experiments with different number of clients for each cell type and report average accuracy of clients holding the same cell type.}
\label{tab:rxrx1result}
\vspace{-1em}
\end{table*}

\begin{figure}[t]
\begin{subfigure}[b]{0.25\textwidth}
    \includegraphics[scale=0.29]{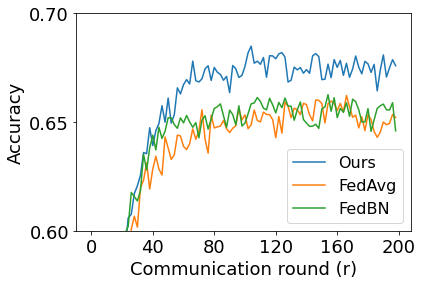} 
    \caption{OfficeHome} 
    \label{fig:OfficeHomeConv}
\end{subfigure}%
\begin{subfigure}[b]{0.25\textwidth}
    \includegraphics[scale=0.29]{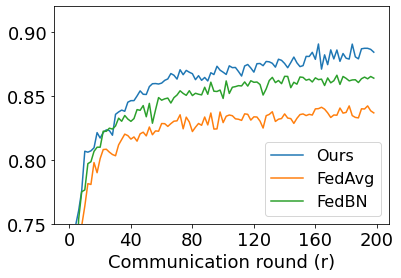} 
    \caption{Digits} 
    \label{fig:DigitsConv}
\end{subfigure}%
\vspace{-0.5em}
\caption{Convergence analysis of FedAvg, FedBN and FRAug on (a) OfficeHome and (b) Digits benchmarks.}
\label{fig:ConvergenceAna}
\vspace{-1em} 
\end{figure}

\subsubsection{Results and Discussion} In \cref{tab:rxrx1result}, we compare FRAug with FedAvg, FedBN, and HarmoFL \cite{jiang2022harmofl}, which is a concurrent work that proposed a strong FL method tailored for heterogeneous medical images, and report the average validation accuracy of clients owning data from the same domain (cell type). We observe that FRAug outperforms all competitors over all settings with different numbers of clients and different data amounts. We highlight the performance improvements achieved by FRAug compared with the baselines, \ie, when 8, 12, and 16 clients join the federated collaborative training, our approach surpasses the other methods by at least 3.04\%, 2.81\%, and 1.21\%, respectively. These results indicate the effectiveness of FRAug on settings with larger quantities of training data as well as its scalability to the complex real-world FL scenarios with more clients.

\subsection{Additional Analyses}
\label{sec:additionalresults}
\subsubsection{Convergence Analysis} 
In \cref{fig:ConvergenceAna}, we display the convergence analysis of the proposed method compared with the baseline FedAvg and FedBN on the OfficeHome and Digits benchmarks. Hereby, we report the average classification accuracy of all clients on their corresponding local testing set after conducting the communication round $r$. As shown in the figure, even though FRAug utilizes the representation augmentation technique, the learning curves of FRAug still exhibit better convergence rates. It's also worth noticing that FRAug already achieves distinct performance gain after 50 communication rounds, \ie, 25\% of the total rounds.

\subsubsection{Analysis of Communication Overhead} 
In \cref{tab:paracompare}, we demonstrate the number of model parameters and computational costs, \ie, the number of operations, of different components used in the proposed method. We observe that both generator and RTNet take only 2-3\% of the parameter numbers used in the classification model, proving the communication overhead between client and server is negligible. Besides, we notice that only less than 1\% of operations are needed for the newly introduced components in FRAug compared with the classification model. Therefore, we conclude that FRAug is communication efficient and does not impose significant impacts on the clients local training, showing its applicability to clients with edge devices and limited computing power.

\begin{table}[t]
\centering
\small
\setlength{\tabcolsep}{5.2mm}
\begin{tabular}{c|c|c}
\toprule
Model & Parameters(M) & MACs(G) \\
\hline
ResNet18 & 11.18 & 1.84 \\
CNN for Digits & 18.15 & 0.08 \\
Generator & 0.39 & $\ll0.01$ \\
RTNet & 0.26 & $\ll0.01$ \\
\bottomrule
\end{tabular}
\vspace{-0.2em}
\caption{Parameters number and MACs (Multiply Accumulate operations) comparison of different components in FRAug.}
\label{tab:paracompare}
\vspace{-1.5em}
\end{table}

\subsubsection{Ablation Study}
\label{sec:ablation}

\begin{table}[t]
\centering
\small
\setlength{\tabcolsep}{2mm}{
\label{tab:ablationoffice}
\setlength\tabcolsep{2.2pt}
\begin{tabular}{c|c|c|ccc}
\toprule
\makecell[c]{G ($\hat{\boldsymbol{v}}$)} & \makecell[c]{RTNet ($\hat{\boldsymbol{u}}$) } & \makecell[c]{EMA 
($\hat{\boldsymbol{u}}_c$) } & \makecell[c]{OfficeHome} & PACS & Digits \\
\hline
& \checkmark & & 64.58{\scriptsize ±0.5} & 88.38{\scriptsize ±0.5} & 86.23{\scriptsize ±0.2}\\

& \checkmark & \checkmark & 65.08{\scriptsize ±0.4} & 88.50{\scriptsize ±0.2} & 86.60{\scriptsize ±0.1}\\ 

\checkmark & & & 65.47{\scriptsize ±0.8} & 90.82{\scriptsize ±0.5} & 87.25{\scriptsize ±0.1}\\ 

\checkmark & & \checkmark & 66.09{\scriptsize ±0.2} & 90.74{\scriptsize ±0.4} & 88.24{\scriptsize ±0.3}\\

\checkmark & \checkmark & & 65.99{\scriptsize ±0.3} & \textbf{91.35}{\scriptsize ±0.1} & 89.51{\scriptsize ±0.1}\\

\checkmark & \checkmark & \checkmark & \textbf{66.60}{\scriptsize ±0.4} & 91.05{\scriptsize ±0.3} & \textbf{89.59}{\scriptsize ±0.2}\\







\bottomrule
\end{tabular}}
\vspace{-0.2em}
\caption{Ablation study for different components of FRAug on three benchmark datasets. The average evaluation accuracy of all clients are reported} \vspace{-1em}
\label{tab:ablationdigits}
\vspace{-0.4em}
\end{table} 

To illustrate the importance of different FRAug components, we conduct an ablation study on three benchmark datasets. The results are shown in \cref{tab:ablationdigits}. We first notice that applying only the client-specific RTNet solely based on local data is ineffective: Its output $\hat{\boldsymbol{u}}^k$ is restricted in the client local distribution when the client-agnostic feature embeddings are inaccessible, which proves the criticality of optimizing a shared generator $G$. We further observe that using the client-agnostic synthetic embeddings $\hat{\boldsymbol{v}}^k$ instead of the personalized versions leads to slight performance gain. This highlights the importance of the transformation by RTNets into personalized client-specific embeddings. Moreover, the results reveal that both types of synthetic embeddings, \ie, $\hat{\boldsymbol{u}}_c^k$ and $\hat{\boldsymbol{u}}^k$, yield a performance boost when used separately. Employing them together further improves the results, which demonstrates their complementarity. 

Additionally, we evaluate the proposed algorithm optimized with different combinations of hyperparameters. From the results, we observe low sensitivity of FRAug to the hyperparameter selection, highlighting its applicability on novel benchmark datasets without time-consuming fine-grained hyperparameter searches. Besides, we conduct experiments with varying numbers of datapoints available on each client. The superior performance of FRAug further indicates its robustness under both data-scarce and data-sufficient scenarios in FL. The detailed evaluation results are provided in Appendix B.

\subsubsection{Robustness to Input Noise} Prior works \cite{jeong2018communication, xu2022acceleration, yoon2021fedmix} focus on generating or adversarially augmenting the clients local training data. On the contrary, the representation generators used in FRAug extract knowledge from the output of the existing feature extractor, i.e., they do not access the input images. More importantly, FRAug does not impose any constraints on the client local update and model aggregation, which indicates its compatibility with the defensive strategies introduced in \cite{enthoven2021overview, lyu2022privacy}.

\begin{table}[ht]
\vspace{-0.5em}
\small
\centering
\setlength{\tabcolsep}{3.5mm}{
\begin{tabular}{c|ccc}
\toprule
Noise Intensity & \makecell[c]{Weak} & \makecell[c]{Medium}  & \makecell[c]{Strong}  \\
\hline
FedAvg & 63.02{\scriptsize ±0.4} & 60.71{\scriptsize ±0.6} & 31.26{\scriptsize ±1.2} \\
FedBN & 63.97{\scriptsize ±0.6} & 60.12{\scriptsize ±0.5} & 30.90{\scriptsize ±0.9} \\
FRAug & \textbf{64.72}{\scriptsize ±1.0} & \textbf{61.45}{\scriptsize ±0.8} & \textbf{31.65}{\scriptsize ±0.7}\\
\bottomrule
\end{tabular}
\vspace{-0.2em}
\caption{Evaluation results of different methods on privatized OfficeHome with different noise intensity. The average accuracy of all clients are reported.}
\label{tab:private}
}
\vspace{-0.5em}
\end{table}

To exhibit the effectiveness of FRAug under the settings with noisy input, we add random noise $\mathbf{\delta}\sim\mathcal{N}(0, \sigma^2\mathbf{I})$ to the client local images when optimizing the classification model. More specifically, we select three noise intensities from weak ($\sigma$ = 0.01), medium ($\sigma$ = 0.1), to strong ($\sigma$ = 1.0). The results in \cref{tab:private} indicate the effectiveness of FRAug under noisy client local data.



\subsubsection{Comparison with Other Augmentation Methods}
\label{sec:augwithrandom}
Since the proposed method applies augmentation in the representation space, we compare FRAug with other augmentation approaches using random noise $\Delta \boldsymbol{u}$ following different distributions. Specifically, we train the prediction head $h$ with real feature embeddings $\boldsymbol{u}$ as well as their augmented variants $\boldsymbol{u}+\Delta \boldsymbol{u}$. We adopt three common distributions for sampling the values of $\Delta \boldsymbol{u}$: Uniform distribution $\mathcal{U}$, Laplace distribution $Lap$ and Gaussian distribution $\mathcal{N}$. We define the standard deviation $\gamma$ of each distribution as a hyperparameter and report the best results. Moreover, we compare our method with concurrent works applying data augmentation, \ie, \emph{FAug} \cite{jeong2018communication} and \emph{FedReg} \cite{xu2022acceleration}.

\begin{table}[t]
\small
\setlength\tabcolsep{3pt}
\begin{tabular}{c|cccc|c}
\toprule
Method & A & C & P & R & avg \\
\hline
FedAvg & 57.47{\scriptsize ±0.6} & 56.74{\scriptsize ±0.9} & 73.32{\scriptsize ±0.8} & 71.25{\scriptsize ±0.3} & 64.69{\scriptsize ±0.6} \\
$\mathcal{U}(-\gamma, \gamma)$ & 56.79{\scriptsize ±0.2} & 57.47{\scriptsize ±0.8} & 72.07{\scriptsize ±0.2} & 73.51{\scriptsize ±0.2} & 64.96{\scriptsize ±0.3} \\
$Lap(0, \gamma)$ & 56.52{\scriptsize ±0.4} & 56.37{\scriptsize ±0.2} & 72.29{\scriptsize ±0.2} & 73.83{\scriptsize ±0.9} & 64.75{\scriptsize ±0.4} \\
$\mathcal{N}(0, \gamma)$ & 56.93{\scriptsize ±0.9} & 57.63{\scriptsize ±0.5} & 72.43{\scriptsize ±0.2} & 73.27{\scriptsize ±0.5} & 65.06{\scriptsize ±0.4} \\
\hline
FAug & 50.18{\scriptsize ±0.5} & 53.48{\scriptsize ±0.9} & 71.82{\scriptsize ±0.4} & 66.08{\scriptsize ±0.8} & 60.39{\scriptsize ±0.7} \\
FedReg & 53.50{\scriptsize ±0.3} & 56.52{\scriptsize ±0.4} & 69.36{\scriptsize ±0.7} & 68.57{\scriptsize ±0.2} & 62.00{\scriptsize ±0.4} \\
FRAug & \textbf{57.61}{\scriptsize ±0.6} & \textbf{60.03}{\scriptsize ±0.5} & \textbf{74.03}{\scriptsize ±0.8} & \textbf{74.58}{\scriptsize ±0.4} & \textbf{66.60}{\scriptsize ±0.3}\\
\bottomrule
\end{tabular}
\vspace{-0.2em}
\caption{Evaluation results of different augmentation methods on OfficeHome benchmark.}
\label{tab:augwithrandom}
\vspace{-1.2em}
\end{table}

In \cref{tab:augwithrandom}, we display the evaluation results of representation augmentation approaches with random noise, as well as the concurrent works, on the OfficeHome benchmark. We notice a distinct performance gap between these methods and FRAug, which further highlights the effectiveness of the proposed method.

\section{Conclusion}
In this work, we present a novel approach to tackle the under-explored feature non-IID problem in FL. The proposed Federated Representation Augmentation (FRAug) method performs client-personalized augmentation in the embedding space to improve the training robustness against feature distribution shift. For that, we optimize a shared generative model to synthesize embeddings by exploiting knowledge from all clients. The output client-agnostic embeddings are then transformed into client-specific embeddings by local Representation Transformation Networks (RTNets). FRAug achieves state-of-the-art results on three benchmark datasets involving feature distribution. Moreover, the superb results of FRAug on a medical dataset illustrate its effectiveness and scalability on complex real-world FL applications.


{\small
\bibliographystyle{ieee_fullname}
\bibliography{main}
}
\end{document}